\documentclass[runningheads]{llncs}
\usepackage{graphicx}
\usepackage{amsmath} 
\usepackage{amssymb}  
\usepackage{subcaption}
\usepackage{booktabs}
\usepackage{bm}
\usepackage{tabularx}
\usepackage{hyperref}
\begin{document}
\title{Towards a Framework for Changing-Contact Robot Manipulation}
%
%
\author{Saif Sidhik\inst{1}\and
Mohan Sridharan\inst{1} \and
Dirk Ruiken\inst{2}}
\authorrunning{S. Sidhik et al.}
%
\institute{Intelligent Robotics Lab, School of Computer Science, \\University of Birmingham, UK\\
        \email{\{sxs1412,m.sridharan\}@bham.ac.uk} \and
Honda Research Institute Europe GmbH, Offenbach am Main, Germany\\
\email{dirk.ruiken@honda-ri.de}}
\maketitle              
%
\begin{abstract}
Many robot manipulation tasks require the robot to make and break contact with objects and surfaces. The dynamics of such \textit{changing-contact} robot manipulation tasks are discontinuous when contact is made or broken, and continuous elsewhere. These discontinuities make it difficult to construct and use a single dynamics model or control strategy for any such task. We present a framework for smooth dynamics and control of such changing-contact manipulation tasks. For any given target motion trajectory, the framework incrementally improves its prediction of when contacts will occur. This prediction and a model relating approach velocity to impact force modify the velocity profile of the motion sequence such that it is $C^\infty$ smooth, and help achieve a desired force on impact. We implement this framework by building on our hybrid force-motion variable impedance controller for continuous contact tasks. We experimentally evaluate our framework in the illustrative context of sliding tasks involving multiple contact changes with transitions between surfaces of different properties.

\end{abstract}
%
%
%
\section{Introduction} 
\label{sec:introduction}
\begin{figure}[t]
\centering
\includegraphics[width=0.6\linewidth]{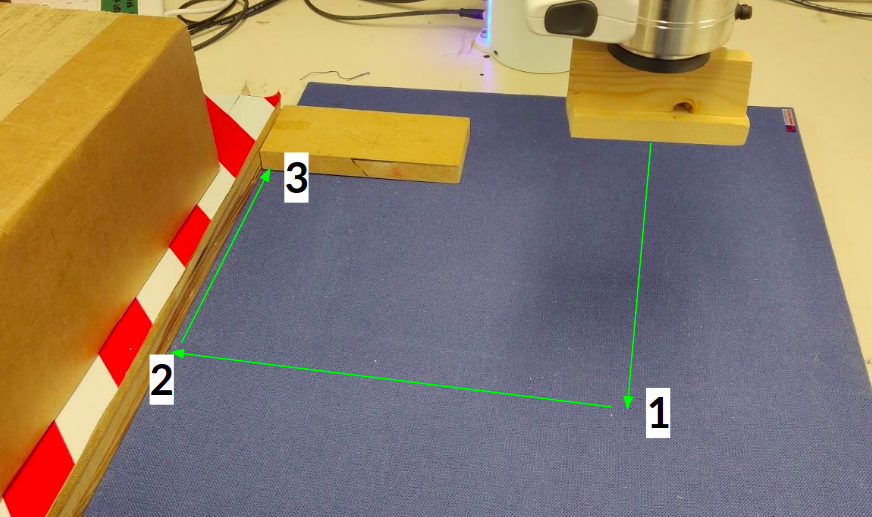}
\caption{Example motion trajectory for a sliding task that involves making and breaking contacts. Specifically the robot has to make contact with the table's surface at ``1" and with another object at ``3".}
\vspace{-1.5em}
\label{fig:task_setup}
\end{figure}

Consider a robot manipulator moving its end-effector along a desired motion pattern (see Figure~\ref{fig:task_setup}), which involves making and breaking contacts, e.g., contact with the table's surface at ``1" and with another object at ``3". This task's \textit{dynamics}, i.e., the relationships between the forces acting on the robot and the resultant accelerations, vary markedly before and after the end-effector comes in contact with the surface. The dynamics also vary based on the type of contact (e.g., surface or edge contact), surface friction, applied force, and other factors. We consider such manipulation tasks involving changes in dynamics due to changes in the nature of contact as ``\emph{changing-contact}'' manipulation tasks. Core industrial assembly tasks, e.g., peg insertion, screwing, stacking, and pushing, and many human manipulation tasks are changing-contact tasks. whose discontinuous dynamics can result in poor transition-phase behavior or instability~\cite{paul1987problems}. Since the interaction dynamics of the robot performing these tasks are discontinuous when a contact is made or broken and continuous elsewhere, it is very difficult to use a single dynamics model or control strategy for these tasks~\cite{brogliato1994transition}. 

Smooth motion along a desired trajectory for a changing contact manipulation task could be achieved using an accurate analytical model of the transitions or a learned model that predicts the transition dynamics. 
Analytical models of the impact dynamics of a system of objects require comprehensive knowledge of the objects' physical and geometric attributes, and often impose unrealistic assumptions not satisfied in practical domains~\cite{hunt1975coefficient,nakashima2015contact}. 
On the other hand, with methods that learn the values of the objects' attributes, build object classifiers based on these attributes, and/or learn sequences of parameters (e.g., joint angles) to achieve the desired trajectory, it is very difficult to acquire sufficient examples of different objects, contacts, features, and trajectories to learn models that can be used to achieve smooth motion without discontinuities~\cite{barragan2014interactive,7174991}.

A different approach for achieving smooth dynamics is to use a \textit{transition-phase} controller that minimizes discontinuities by lowering velocity and stiffness to reduce impact forces, vibration, and jerk on impact. Existing transition control strategies switch to a different controller once a contact is detected, but this switch could still cause substantial discontinuities in the interaction dynamics, damaging the robot or the objects~\cite{mills1993control,sidhik_learning}. We instead seek to predict contacts accurately and adapt the velocity and stiffness during the transition phase to minimize discontinuities, allowing the robot to switch to a different controller after contact. To do so, we need to determine: (\textbf{Q1}) how best to predict when contact will occur? (\textbf{Q2}) when to activate the transition-phase controller? and (\textbf{Q3}) how best to adapt the transition-phase controller's parameters to the task? We also need to determine: (\textbf{Q4}) what representation and control strategy to use for reliable and efficient control with minimal samples before, during, and after contact? Our framework builds on our prior work on a task-space, hybrid force-motion, variable impedance controller for continuous contact tasks~\cite{sidhik_learning}, which partially addresses Q4 and simplifies the problem, enabling the following as solutions to questions stated above: 
\begin{itemize}
    \item A simple and efficient contact prediction method that incrementally improves its estimates.
    \item An adaptive strategy that uses predicted contacts to minimize time spent in the transition-phase;
    \item An approach that revises the velocity to be used in the transition phase to achieve a  $C^\infty$ smooth velocity profile and a desired impact force. 
\end{itemize}
We evaluate our framework on a physical robot and in simulation, using the motivating example of sliding tasks that involve making and breaking contacts with objects and surfaces of different attributes. To thoroughly explore the control problems, we only consider sensor input from a force-torque (FT) sensor in the wrist. We review related work (Section~\ref{sec:relwork}), and describe the framework (Sections~\ref{sec:arch-overview}-~\ref{sec:ctrl}), experimental results (Section~\ref{sec:experiments}), and conclusions (Section~\ref{sec:discuss}).

\section{Related Work}
\label{sec:relwork}
As stated earlier, control for changing contact manipulation tasks can be achieved using analytical methods, learning methods, or transition-phase controllers.

Analytical methods that explore the relation between relative motion of two colliding objects and their impact dynamics mainly typically formulate it as a linear complementary problem (LCP) that considers the velocities and impulses~\cite{pfeiffer1996multibody}, or the accelerations and forces~\cite{yashima2003complementarity}, at contact points~\cite{hunt1975coefficient,moore1988collision,routh1955dynamics,nakashima2015contact}. 
Such a formulation and the associated methods can guarantee physical consistency between motion and impulses, but it is computationally expensive to solve the LCP at every time step, and difficult to provide the required prior knowledge of object attributes and/or accurate 3D object models in complex domains.

Methods developed to learn the physical attributes of objects, or to categorize objects based on these attributes, require the robot to perform the related task multiple times to obtain the training examples needed for building relevant models or optimizing the models' parameters~\cite{7174991,barragan2014interactive}. Also, the learned models need to be retrained if the objects, tasks, or interaction dynamics change over time. 

Methods that use a transition-phase controller (for changing contact manipulation tasks) focus on minimizing the discontinuities in the dynamics, i.e., on reducing the forces, vibration, and jerk on impact~\cite{mills1993control,sidhik_learning}. However, many of these methods switch to a different controller only after a contact is detected, which can result in significant discontinuities when the switch is made, along with loss of energy, and damage to the robot or the domain objects.

In this paper, we pursue an approach based on a transition-phase controller. However, instead of abruptly transitioning to a controller after contact has been detected, we focus on anticipating contacts and adapting the velocity and stiffness used in the transition phase to reduce impact forces, vibration, and jerk on impact. To ensure that the resultant motion is smooth, the transition velocity profile has to be continuous; to make the motion smooth in acceleration and jerk, the motion needs to be at least $C^4$ smooth. Methods have been developed for kinematic time-optimal motion using trapezoidal velocity profiles~\cite{grassmann2018smooth}, $C^4$ smooth trajectories using multiple trajectory segments~\cite{nam2004study}, and for minimum-jerk motion profiles~\cite{piazzi2000global,freeman2012minimum,huang2006global}. However, these methods are computationally expensive when the robot has to compute a path through many points in each trajectory segment. Our approach instead seeks to provide smooth motion over any given trajectory, modifying the trajectory to transition to a desired velocity using a velocity profile that is $C^\infty$ smooth. This is achieved by making suitable representational choices, accurately predicting contact, and suitably transitioning into (out of) the transition phase controller before (after) contact. The main contribution of this work is a framework that can, in a few trials, learn to (a) predict contact changes in a changing-contact manipulation task; and (b) adapt its control strategy before impact such that the impact forces are reduced while deviating from the provided trajectory as little as possible.

\section{Framework Overview}
\label{sec:arch-overview}
\begin{figure}[tbh!]
\vspace{0.8em}
\centering
\includegraphics[width=0.8\linewidth]{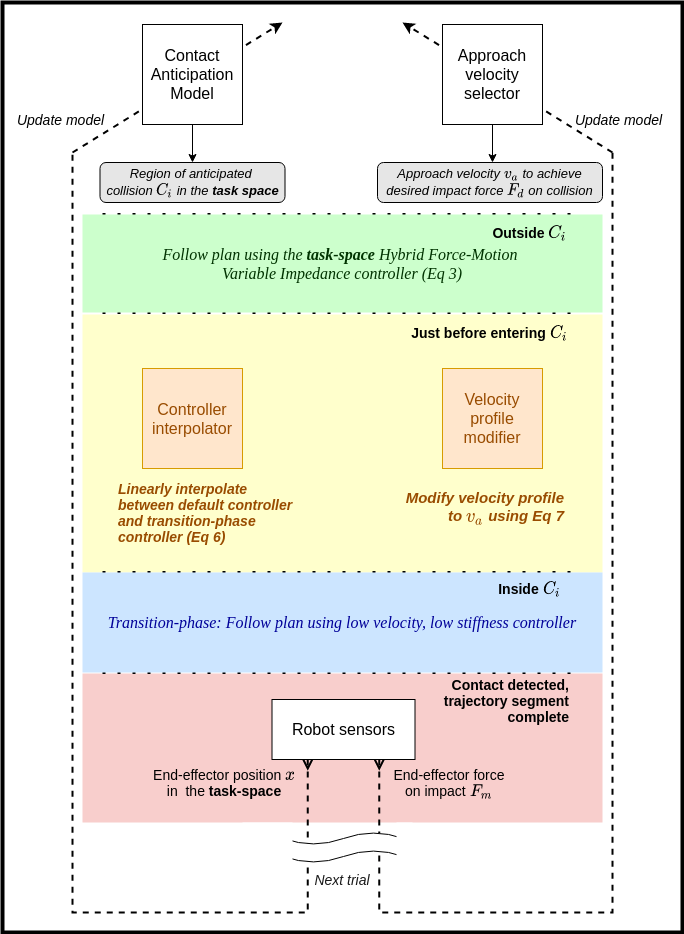}
\caption{Overview of our framework for smooth control of changing-control manipulation tasks.}
\vspace{-1em}
\label{fig:framework-diagram}
\end{figure}

Figure~\ref{fig:framework-diagram} presents an overview of our framework. The inputs are the desired motion trajectory, the force-torque sensor measurements, and the end effector position. The default controller is the hybrid force-motion variable impedance controller that we developed for continuous contact tasks~\cite{sidhik_learning}. This includes an incrementally learned forward (predictive) model of end effector measurements; the error between the predicted and actual measurements automatically revise stiffness values in control laws to determine the control signal. We showed that operating in task (i.e., Cartesian) space allows this controller to use suitable abstractions to learn accurate forward models from very few examples, provide compliance along specific directions, and accurately track the desired trajectory, thus partially addressing Q4 in Section~\ref{sec:introduction}. The framework in this paper builds on this default controller's representation for changing-contact manipulation. Specifically, we introduce a task-space contact anticipation model that incrementally updates its contact prediction using a Kalman filter (Q1). These predictions are used to minimize the time spent in the transition phase (Q2), and the velocity to be used in the transition phase is set adaptively to achieve a $C^\infty$ smooth velocity profile and a desired impact force (Q3). Once the transition is completed at a suitable velocity and stiffness to minimize discontinuities, the robot moves to using another version of the default controller and revises the parameter values automatically as needed. We begin with a description of the contact prediction method.

\section{Contact Prediction}
\label{sec:contact-predict}
Anticipating contacts by predicting impact forces or time to collision is challenging because these parameters are influenced by robot dynamics and the controller's parameters. For instance, reducing the robot's velocity or its controller's stiffness reduces the impact force and increases the time to contact. 
It is more reliable to predict \textit{static} contact parameters such as \textit{end-effector pose during impact} and \textit{direction of contact force}, which do not change significantly for repetitions of the task as long as we can assume that the trajectory and the environment's attributes do not change significantly; these are reasonable requirements for many tasks.

The robot's belief about the position of each expected collision while executing the assigned motion trajectory is modeled as a multivariate Gaussian in the workspace, with the covariance ellipsoid denoting the uncertainty along different motion control dimensions; it is also the ``region of anticipated contact'' $\mathcal{C}$.
Given that the controller operates in the task space, each contact location's representation is compact and is updated over very few trials of the task using a Kalman filter with the state update equation: $\mathbf{\dot{x}} = \mathbf{A}\mathbf{x}_k + \mathbf{B}\textbf{u}_k + w$, where $\mathbf{x}$ is the contact position, $\mathbf{A}$ is the object's self-activation, i.e., it defines the dynamics of motion of the contact point without the robot acting on it ($I$ for positively activated objects, which is the case in this paper), $\mathbf{B}$ is the control matrix capturing the effect of the robot's action $\textbf{u}$ on contact position, and $w$ is Gaussian noise modeling the uncertainty in the contact location.
The sensor model uses the end-effector pose (as given by forward kinematics from joint positions) as measurement when a contact is detected. This sensor model (with noise depending on the joint encoder
noise and forward kinematics) will provide a corrected estimate of the contact point once a contact is made, resulting in a reduced covariance ellipsoid for subsequent trials.
The Kalman filter's update equations are as follows:
\begin{subequations}
\begin{align}
  &\mathbf{\hat{\mu}}_{k|k-1} = \mathbf{A}_{k-1} \mathbf{\hat{\mu}}_{k-1|k-1} + \mathbf{B}_{k-1}\mathbf{u}_{k-1} \label{eq:kf_pred1}\\
  &\mathbf{\Sigma}_{k|k-1} = \mathbf{A}_{k-1}\mathbf{\Sigma}_{k-1|k-1}\mathbf{A}_{k-1}^\mathbf{T} + \mathbf{Q}_{k-1}\label{eq:kf_pred2}\\
  &\mathbf{v}_k = \mathbf{y}_k - \mathbf{H}_k \mathbf{\hat{\mu}}_{k|k-1}\label{eq:kf_update1}\\
  &\mathbf{S}_k = \mathbf{H}_k\mathbf{\Sigma}_{k|k-1}\mathbf{H}_k^\mathbf{T} + \mathbf{R}_k \\
  &\mathbf{K}_k = \mathbf{\Sigma}_{k|k-1}\mathbf{H}_k^\mathbf{T}\mathbf{S}_k^{-1}\\
  &\mathbf{\hat{\mu}}_{k|k} = \mathbf{\hat{\mu}}_{k|k-1} + \mathbf{K}_k\mathbf{v}_k\\
  &\mathbf{\Sigma}_{k|k} = \mathbf{\Sigma}_{k|k-1} - \mathbf{K}_k\mathbf{S}_k\mathbf{K}_k^\mathbf{T}\label{eq:kf_update2}
\end{align}
\end{subequations}
where $\mathbf{\hat{\mu}}_{i|i-1}$ and $\mathbf{\Sigma}_{i|i-1}$ are the predicted mean and covariance at step $i$, $\mathbf{\hat{\mu}}_{i|i}$ and $\mathbf{\Sigma}_{i|i}$ are the corrected mean and covariance based on measurement $\mathbf{y}_i$ (of pose on contact) at step $i$, $\mathbf{K}$ is the Kalman gain, and $\mathbf{Q}$ and $\mathbf{R}$ are noise matrices. Our representational choices enable us to develop a contact anticipation model that is simple, efficient, and reliable; a linear model is used to accurately estimate contact locations in the workspace from very few (noisy) repetitions of the task.
Although this representation supports contact with movable objects, we assume (in this paper) that the end effector only makes contact with stationary objects (i.e., $\mathbf{A} =\mathbf{I}$, $\mathbf{B} = \mathbf{0}$). Also, $\mathbf{H} = \mathbf{I}$ since state and measurements are in the same space. These simplifications reduce the Kalman updates to simple Gaussian updates using the noisy measurements based on the robot's kinematics model (since sensor input is limited to that from an FT sensor) each time the robot experiences a particular contact change.


\section{Controller Formulation}
\label{sec:ctrl}
For any given task, the desired motion trajectory $P$ is provided as a sequence of mappings from time to the target end-effector pose and force (for force control). This is typically obtained through a single demonstration of the task, with the human moving the robot manipulator. As stated earlier, the controller is defined in the Cartesian-space; it provides an intuitive trajectory description and any obstacles can be considered efficiently.
Also, $P$ is assumed to be made up of segments of smooth and continuous pose trajectories that satisfy the minimum jerk constraint. The motion within each segment is smooth, but each segment ends with a change in the direction of control, and $P$ does not account for the collisions (i.e., contact points).

Our framework's default controller builds on a standard variable impedance control equation~\cite{villani2008handbook}:
\begin{equation}
    \bm{h}_c = \bm{\Lambda} (\bm{q}) \ddot{\bm{x}}_d  + \bm{\Gamma}(\bm{q,\dot{q}})\dot{\bm{x}}_d + \bm{\eta}(\bm{q}) + \mathbf{K_p}\Delta\bm{x} + \mathbf{K_d}\Delta\bm{\dot{x}} \label{eq:ctrl_law}
\end{equation}
where $\bm{h}_c$ is the task-space control command, $\bm{\Lambda}
(\bm{q}) = (\bm{J}\bm{M}(\bm{q})^{-1}\bm{J}^T)^{-1}$ is the $6\times
6$ operational space inertia matrix, $\bm{\Gamma}(\bm{q,\dot{q}}) =
\bm{J}^{-T}\bm{C}(\bm{q,\dot{q}})\bm{J}^{-1} - \bm{\Lambda}
(\bm{q})\dot{\bm{J}}\bm{J}^{-1}$ is the compensation wrenches
including centrifugal and Coriolis effect, and $\bm{\eta}(\bm{q}) =
\bm{J}^{-T}\bm{g}(\bm{q})$ is the gravitational wrench.
$\bm{M}(\bm{q}), \bm{C}(\bm{q},\bm{\dot{q}})$ and $\bm{g}(\bm{q})$ are
the equivalent values defined in the joint space of the robot;
$\Delta\bm{x}$ is the error in end-effector pose with respect to a
desire pose $\bm{x}_d$; and $\mathbf{K_p}$ and $\mathbf{K_d}$ are $6\times
6$ symmetric positive-definite matrices defining the desired impedance
stiffness and damping. The final joint-space torque control is
computed as $u = \bm{J}^T h_c$.



 
In the absence of the external wrench $\bm{h}_{e}$, the control law
provides asymptotic stability with equilibrium state $\bm{\dot{x}}_e =
0, \Delta\bm{x} = 0$ for a closed-loop system. In the presence of
non-zero $\bm{h}_{e}$, a non-null $\Delta\bm{x}$ will be present at
equilibrium. For a fixed or non-stationary target $\bm{x}_d$, if the
external force $\bm{h}_{e}$ is known to be due to non-fixed
resistance (e.g., friction when sliding on a surface), forces against
the direction of motion can be cancelled out with an appropriate
feed-forward term $\bm{h}_{f\!f}$ in the control law:
\begin{equation}
    \bm{h}_c = \bm{\Lambda} (\bm{q}) \ddot{\bm{x}}_d  + \bm{\Gamma}(\bm{q,\dot{q}})\dot{\bm{x}}_d + \bm{\eta}(\bm{q}) + \mathbf{K_p}\Delta\bm{x} + \mathbf{K_d}\Delta\bm{\dot{x}} + \bm{h}_{f\!f}\label{eq:final_ctrl_law}
\end{equation}
When the robot is not expecting a contact change, $\bm{h}_{f\!f}$ and the impedance gains ($\mathbf{K_p}$, $\mathbf{K_d}$) are revised based on the difference between the predicted and observed values of forces and torques at the end effector. The predictions are based on a task-specific (feed)forward model that is revised incrementally during task execution~\cite{sidhik_learning}. We build on this default controller to answer questions Q2-Q4.

\subsection{Transition Controller Parameters}
We first describe the formulation of the transition controller and its parameters. Since the permitted impact force may differ based on the task, e.g., large forces can damage delicate objects in certain tasks, it is reasonable to assume a safe limit on the maximum allowed impact force.
Also, experimental analysis indicated that reducing the controller stiffness helps reduce the jerk in motion after impact by providing compliance, but has no significant effect on impact forces because the robot has to make the contact for the error and stiffness term in the feedback control loop to come into effect. A safe controller should thus have lower stiffness for reducing vibrations. In addition, the approach velocity was observed to be directly proportional to the force at impact, especially when the robot registers a contact while moving in free space. One advantage of the design of our controller (and the related representational choices) is that a simple approach (linear regression) can be used to fit the relationship between impact force and the approach velocity between a pair of objects. This relationship is then used to compute the approach velocity for a desired impact force. 



The robot may not have a model of these relationships when performing a task for the first time.
So, for any given target impact force, the robot starts with a safe low velocity during the first trial, using the difference between the target impact force and the measured impact force to revise the approach velocity for the next iteration of the task:
\begin{equation}
    \Delta v_{a} = \beta (F_{d} - F_{m})  \label{eq:approach_vel_update}
\end{equation}
where $\Delta v_a$ is the value used to revise the approach velocity, $F_d$ is the desired impact force along motion direction, $F_m$ is the measured impact force, and $\beta$ is a learning rate that is ideally a value less than or equal to the slope of the plot relating impact force to the approach velocity. Over time, this approach enables the robot to learn a task-specific velocity of approach for a desired impact force. The learned linear model can also be reused for other target impact forces. Given this formulation, we next describe our approach to decide when to switch to using the transition-phase controller.

\subsection{Switching to Transition-Phase Controller}
\label{sec:boundary}
Recall that a lower stiffness in the transition phase can reduce vibrations on contact, and a lower velocity reduces the impact forces. Since any such strategy will cause the robot to deviate from the desired trajectory,
the robot should ideally switch to this control phase just before the
contact is made, and switch out of it immediately after stable contact
is established. Since this is not possible in practice, it is safer to
switch to this control mode when it enters a region in the task space
where the contact is highly likely to occur, and switch out of it
once stable contact is achieved.

As stated in Section~\ref{sec:contact-predict}, we use the covariance of the multivariate Gaussian estimating the contact location to define the region of anticipated contact $\mathcal{C}$ in the
task-space. Activating the transition-phase controller just before or after it enters $\mathcal{C}$ ensures that the transition-phase is only active when a contact/collision is anticipated. The part of the target motion trajectory $P$ within $\mathcal{C}$ can be found bychecking if the points in $P(t)$ satisfy the relation given by:
\begin{equation}
  (\mathbf{P}(t)-\mathbf{\mu})^T \mathbf{\Sigma}^{-1}(\mathbf{P}(t)-\mathbf{\mu}) \leq \lambda \label{eq:ell}
\end{equation}
where $\mathbf{\mu}$ is the mean of the Gaussian predicting contact position,
$\mathbf{\Sigma}$ is the covariance, $\lambda$ is the scaling factor governed by the confidence in the covariance estimate; it is modeled as the chi-squared percent point function of the desired confidence value.
The first point in trajectory $P$ to satisfy this condition is the boundary $p_c$ of the anticipated collision region $\mathcal{C}$. When the robot does a task for the first time, the position uncertainty and hence the volume of $\mathcal{C}$ are large, and the robot switches to the transition phase controller (with lower stiffness and velocity) earlier than actual impact. Over time, as the covariance ellipsoid shrinks, the robot's contact prediction is more accurate and it transitions to the transition controller when it is just about to make impact.

\subsection{Smooth Transition between Controllers}
Since the desired impact force is primarily achieved by revising the approach velocity, the transition-phase controller is set (by the designer)to use lower fixed controller gains $(\mathbf{K_{p}}^*, \mathbf{K_{d}}^*)$ as the robot moves at a lower velocity, for reducing the negative effects of collision. To avoid discontinuities, the robot needs to smoothly transition from a normal (pre-contact) controller with output $\mathbf{u_1}$ to the transition-phase controller with output $\mathbf{u_2}$. We use linear interpolation of $\mathbf{u_1}$ and $\mathbf{u_2}$ over a time window $[0,T]$ such that the transition is completed by the time the robot reaches $p_c$:
\begin{align}
  \mathbf{u} = &(1 - \alpha)\mathbf{u_1} + \alpha \mathbf{u_2};\qquad
  \alpha = t/T \qquad t \in [0,T]\label{eq:ctrl_interp}
\end{align}
where $T$ is the desired duration of the transition between the controllers. As long as the outputs from the two controllers ($\mathbf{u_1}$ and $\mathbf{u_2}$) are individually smooth, the output of the combination will also be smooth. In this work, controllers use the task-space representation described earlier, with $\mathbf{u_2}$ being the output of the fixed, low-gain, transition-phase controller as the arm approaches the contact point. A similar approach is used to smoothly transition from the transition-phase controller to a normal controller after contact is made.


\subsection{Modifying Velocity Profile over Target Trajectory}
\label{sec:velocity-profile}
Transition-phase controllers typically use a lower velocity than the
original kinematic sequence $P$ to reduce the force at impact. Also,
as the region $\mathcal{C}$ is revised by the Kalman filter, the robot will switch to the
desired approach velocity at different points in the target trajectory in
different trials. Therefore, the timeline of the trajectory has to
be modified to account for the modified velocity profile.

To modify the given motion trajectory such that velocity changes
smoothly, we enable the robot to create a new velocity profile and
time-mapping.
Our approach builds on the trapezoidal formulation used in literature for velocity profiles; it can be viewed as either the \textit{lift-off} or \textit{set-down} phase of a trapezoidal profile. Unlike other representations, our formulation results in motion that is smooth and continuous at all orders, i.e., is $C^\infty$ smooth.

Without loss of generality, assume that the original motion trajectory $P$ is along one dimension with velocity $v_1$. Assuming that transition starts at time $t_1$ with $v_1$ and has to be completed at $t_2$ (i.e., in $t_2 - t_1$) with approach velocity $v_2$ (as the robot crosses boundary point $p_c$ of $\mathcal{C}$), the velocity profile is:
\begin{equation} v(\tau) = \begin{cases}
    v_1 + \frac{(v_2 - v_1) e^{-1/\tau}}{e^{-1/\tau} + e^{-1/(1-\tau)}} &\quad\text{if }0 < \tau < 1,\\
    v_1 & \quad\text{if }\tau \leq 0\\
    v_2 & \quad\text{if }\tau \ge 1\\
  \end{cases} \label{eq:vel_prof}
\end{equation}
where $\tau = t/T = t/(t_2 - t_1)$. When $0 < \tau < 1$, $e^{-1/\tau}$ has continuous derivatives of all orders at every point $\tau$ on the real line. Since $v(\tau)$ has a strictly positive denominator for all points on the real line, it is smooth. Due to the enforcement of velocity limits when $\tau$ is outside $[0, 1]$, the velocity profile provides a smooth transition from $v_1$ to velocity $v_2$ in interval $[t_1, t_2]$. Although defined as being piece-wise, the function is continuous along the real line.
 
 \begin{figure}[!ht]
 \vspace{-1em}
  \begin{center}
    \includegraphics[trim={0cm 0cm 0 0.7cm},clip,width=\linewidth]{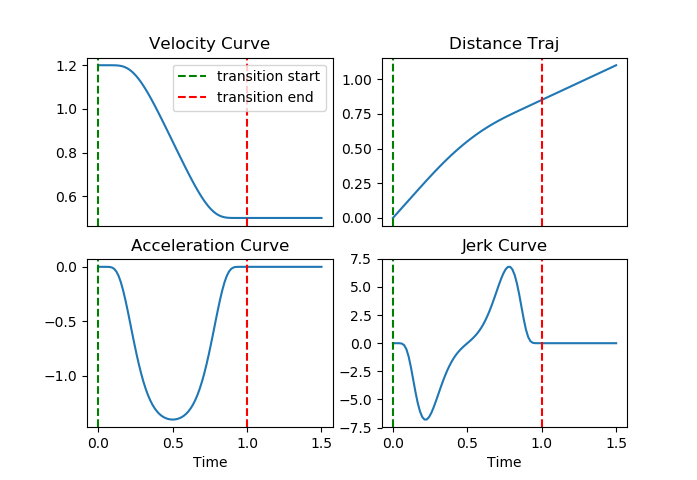}
  \end{center}
  \vspace{-2em}
  \caption{Velocity plots with matched position, acceleration, and jerk plots. Velocity varies from 1.2 to 0.5 in unit time.}
  \label{fig:velprofeg}
  \vspace{-1.em}
\end{figure}
The velocity profile and its associated position, acceleration and
jerk plots are shown in Figure~\ref{fig:velprofeg}. The position
trajectory is obtained by integrating the velocity profile with
respect to $\tau$. All motion derivatives are
continuous throughout the transition.
The timeline of the new velocity profile $v(\tau)$ for any given contact) can be used to modify the target trajectory $P$ such that the velocity transition is completed as the robot reaches $p_c$. As the velocity transition duration (Equation~\ref{eq:vel_prof}) and the controller transition duration (Equation~\ref{eq:ctrl_interp}) are the same, the robot will become compliant and slow down just before it enters $\mathcal{C}$.

Note that only the position is adapted according to the $C^\infty$ profile; the orientation profile is modified using SLERP, a linear interpolation of points in the spherical space of quaternions. This approach may cause position-orientation mismatch in complex manipulation tasks involving significant orientation changes at transition regions.

\section{Experimental Analysis}
\label{sec:experiments}
We experimentally evaluated the following hypotheses: 
\begin{itemize}
\item[\textbf{H1:}] Our contact prediction approach accurately predicts contact pose and improves estimate over time, reducing the task-completion delay and trajectory tracking error;

\item[\textbf{H2:}] A learned relation between approach velocity and the impact force provides an accurate estimate of the approach velocity for a desired impact force; and

\item[\textbf{H3:}] Overall framework produces smooth motion dynamics (velocity, acceleration, etc.) throughout the manipulation task with multiple contact changes.
\end{itemize}
For experiments, we used a 7-DoF Franka Emika Panda robot operating on a tabletop (Figure~\ref{fig:task_setup}) and its simulated version in PyBullet. \textit{We focus on describing and discussing the results obtained on the physical robot platform in this paper; a video of the physical robot and additional simulation results are in the supplementary material\footnote[1]{\url{https://drive.google.com/drive/folders/1mN0r5Gi37TT4goHmumt029rz0nhZ5Wdd?usp=sharing}}}.
The evaluation measures include position tracking accuracy, task completion time, and the time spent in the transition-phase.

\subsection{Contact Anticipation}
To evaluate the contact prediction ability (\textbf{H1}), we used a task-space trajectory that required the robot to approach a (static) table from above, move back up without making contact, and move down and make contact with the surface, resulting in a zig-zag trajectory along the z-axis. The initial guess of the contact position was provided manually to simulate input from an external planner or vision system. The robot should ideally be compliant and moving with a lower velocity when approaching a contact point, but it should spend as little time as possible in this low-velocity, low-stiffness transition phase to reduce tracking error and delay in task completion. This desired behavior is dependent on the robot's ability to incrementally and quickly improve its belief about the location of each contact position in the task space.

The initial state of the predictor is provided manually and is associated with a larger covariance (0.175 along each of the three dimensions, with distance measured in meters) emulating the uncertainty associated with a noisy vision sensor or planner. The target motion trajectory in conjunction with the large covariance caused the robot's region of anticipated contact ($\mathcal{C}$) to overlap with points in the first `valley' of the zig-zag pattern when there is no actual contact with the table's surface. If the contact prediction improves over time, we expect the robot to obtain a tighter estimate of $\mathcal{C}$ over time. The robot should then not switch to the transition-phase controller in the first valley, but do so when it approaches the table the second time. Given the focus on contact prediction, we empirically chose safe values for the transition-phase control parameters (i.e., approach velocity and stiffness).

\begin{figure}[!ht]
\centering
\includegraphics[width=\linewidth]{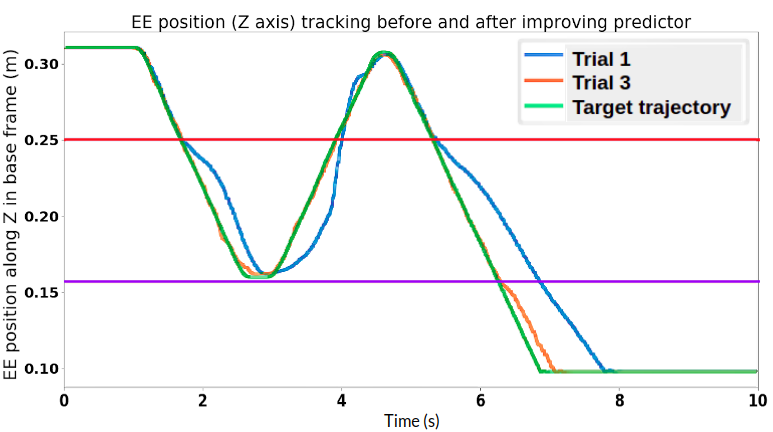}
\vspace{-1.5em}
\caption{Tracking position of end-effector (EE) during trial-1 and trial-3 of contact prediction with Kalman filter. The red horizontal line is the edge of the covariance ellipsoid in trial 1; the violet line is the ellipsoid boundary in trial 3. Updated covariance in trial 3 enables the robot to avoid going to the transition-phase in the first dip of zig-zag trajectory and reduce the tracking error.}
\vspace{-0.5em}
\label{fig:kftrack}
\end{figure}

We observed a significant reduction in covariance, e.g., from $0.175$ to $0.07$ in just three iterations in a set of trials summarized in Figure~\ref{fig:kftrack}, which enabled the robot to avoid going to the transition-phase in the first dip in the trajectory. Also, the average Euclidean position tracking error of the end-effector (EE) per time step reduced from $1.3\,cm$ in the first trial to $0.16\,cm$ in the third trial, and the task completion time reduced from $7.9\,s$ in the first trial to $7.2\,s$ in the third trial; the expected (ground truth) motion duration is $7\,s$. Similar results were obtained with other target trajectories, indicating support for \textbf{H1}, i.e., that the contact prediction quickly improves the belief about the contact position, reducing delays in task completion as well as errors in trajectory tracking. These results also indicate that using the transition-phase controller only when it is required reduces the deviation from the desired motion trajectory.


\subsection{Approach Velocity and Impact Force}\label{sec:vel_impact_expt}
\begin{figure}[!ht]
\vspace{0.8em}
\centering
\includegraphics[width=\linewidth]{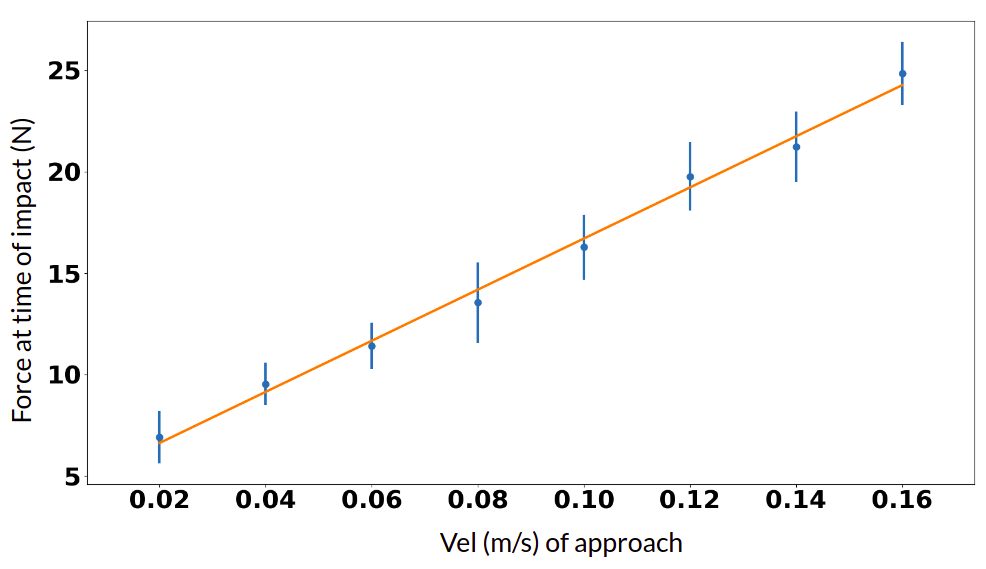}
\vspace{-2em}
\caption{Approach velocity vs force on impact. Orange line denotes the estimated linear relationship.}
\vspace{-1em}
\label{fig:vel_imp_lin}
\end{figure}

To test the relation between approach velocity and impact force on contact, the robot was given a target motion trajectory involving moving in free space and making contact with table; this is also shown in the supplementary video. 
The task was repeated with different velocities ranging from $0.02\,m/s$ to $0.16\,m/s$ in steps of $0.02$, each repeated four times, and we measured the corresponding force on contact. We observed that a line whose parameters were estimated by linear regression provided a reasonably good fit for the relationship between end-effector approach velocity and the end-effector force along the direction of motion; see Figure~\ref{fig:vel_imp_lin}. The variance in the fit can be attributed largely to the noise in the force-torque sensor, which is significantly troublesome during discontinuities such as collisions.

Given such a learned relationship, the robot was asked to perform the same target trajectory (as above) but choose its approach velocity so as to achieve a given impact force on contact. The ground truth force measured after contact was compared with the desired impact force. Table~\ref{tab:vel_est_imp_free_space} summarizes results for four trials for four of the $11$ target force values we tested ($10-20$N at $1$N increments). We observed that the robot was able to compute an approach velocity that resulted in an impact force similar to the desired value, with an error of $\sim3\,$N. 
These errors were more likely at lower values of the target impact force and can largely be attributed to sensor noise, i.e., the learned model is limited by the accuracy, sensitivity, and resolution of the force-torque sensor, joint encoders, and the robot's forward kinematics model.

\begin{table}[tb]
\vspace{1em}
\centering
\begin{tabular}{lccr}
\toprule
 \begin{tabular}[c]{@{}l@{}}Target \\ Force (N)\end{tabular} & \begin{tabular}[c]{@{}l@{}}Estimated reqd\\velocity (m/s)\end{tabular} & \begin{tabular}[c]{@{}l@{}}Measured\\force (N)\end{tabular} & \begin{tabular}[r]{@{}r@{}}Error\\force (N)\end{tabular} \\
\midrule
10               & 0.047                                                                       & 7.4                                                           & 2.6             \\ 
12               & 0.063                                                                       & 15.1                                                          & 3.1             \\ 
15               & 0.086                                                                       & 15.3                                                          & 0.3             \\ 
18               & 0.11                                                                        & 16.7                                                          & 1.3             \\ 
\bottomrule
\end{tabular}
\caption{Linear relationship can be used to specify approach velocity as a function of the desired impact force. Errors observed predominantly at lower values of target force due to sensor noise.}
\label{tab:vel_est_imp_free_space}
\vspace{-1.5em}
\end{table}

\begin{figure}[tb]
\centering
\includegraphics[width=\linewidth]{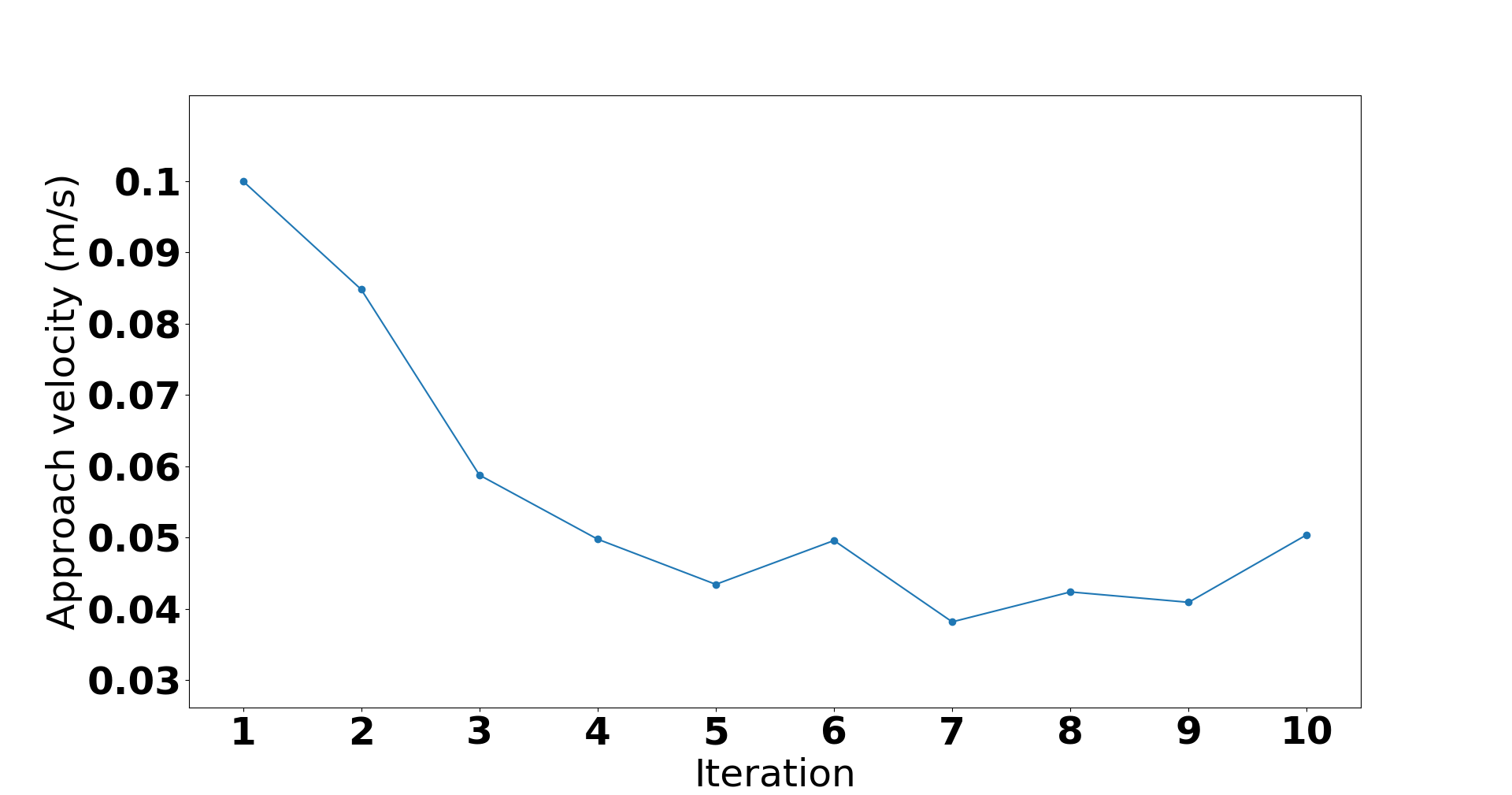}
\vspace{-1.5em}
\caption{Approach velocity adaptation over different trials to achieve target impact force of $10\,$N.}
\vspace{-1em}
\label{fig:vel_converge}
\end{figure}

The velocity update rule based on Equation~\ref{eq:approach_vel_update} was then tested on the robot without providing the previous linear model. The initial value of the approach velocity was set to $0.1\,m/s$, the target impact force $F_d = 10$\,N, and $\beta =0.003$. Figure \ref{fig:vel_converge} shows the evolution of approach velocity across 10 trials. It can be seen that the velocity of approach remains roughly around $0.045\,m/s$ from the fifth iteration of the task, after which the noise of force torque sensor at impact makes further convergence difficult. The error between measured and desired force reduced from 8.5\,N to 0.2\,N at the end of 10 iterations. Similar results were obtained for other values of initial approach velocity and target impact force. These results indicate that in the absence of the learned linear model, it takes the robot a greater number of iterations (and a suitable choice of initial velocity) to compute a suitable approach velocity for a target impact force. These results thus indicate support for \textbf{H2}.

\subsection{Smoothness of Motion}

The motion profiles (i.e., velocity, acceleration, etc.) of a changing-contact manipulation task are expected to have large spikes in the absence of a framework for predicting the contact locations and adapting the velocity and stiffness during the approach to a contact position. This hypothesis was tested in a simulated environment which showed that using the framework significantly reduces the spikes in the overall motion profile of the robot in a task, while also ensuring safe interaction during contact changes. The results from these simulation experiments can be found in the supplementary material, and are not included in this paper due to space limitations.

\begin{figure*}[!ht]
  \begin{center}
    \begin{subfigure}{\textwidth}
      \includegraphics[width=\textwidth]{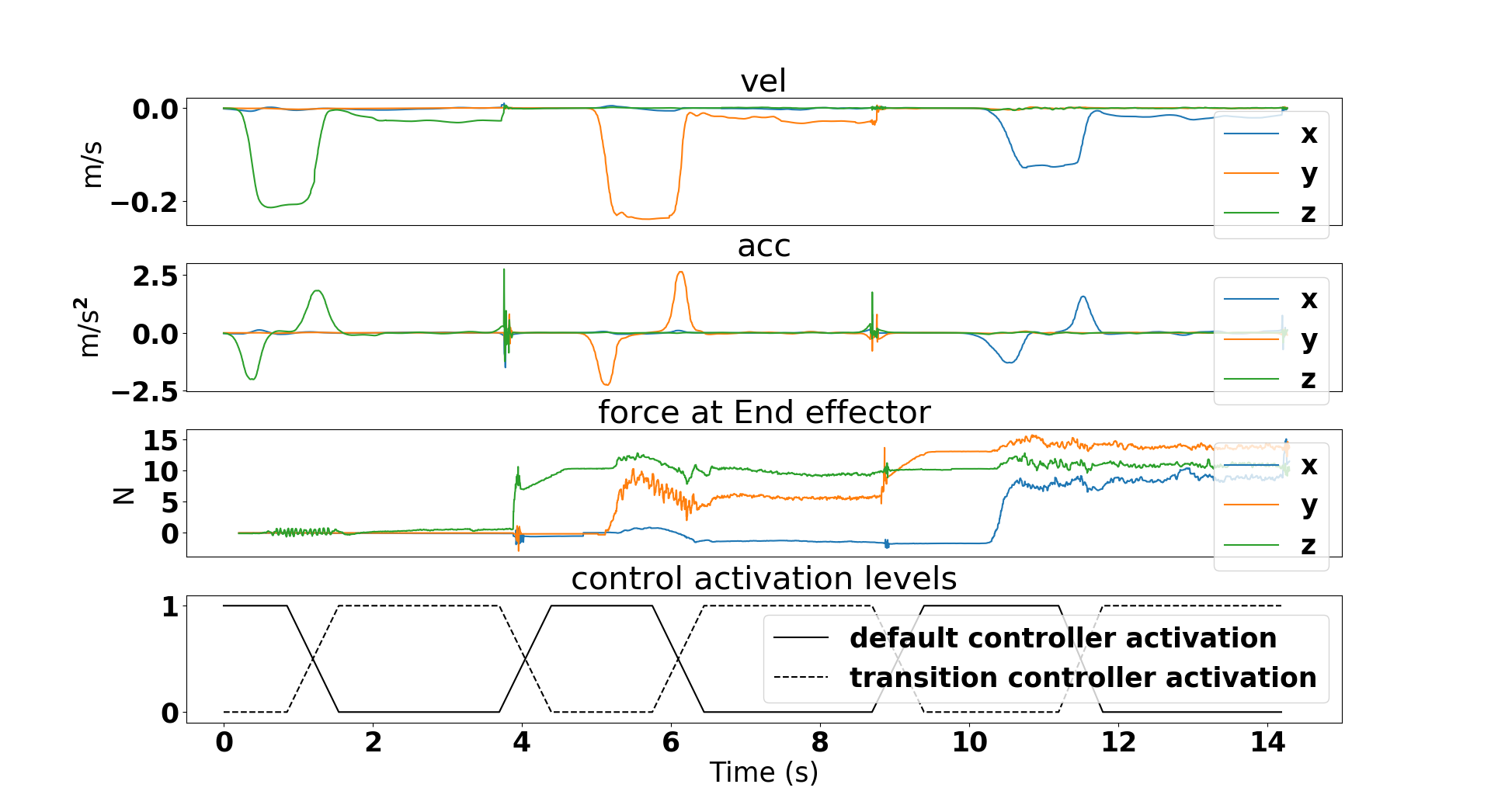}
      \vspace{-1em}
      \caption{Experiment trial 1.}
      \label{fig:expt_first_trial}
    \end{subfigure}
    \begin{subfigure}{\textwidth}
      \includegraphics[width=\textwidth]{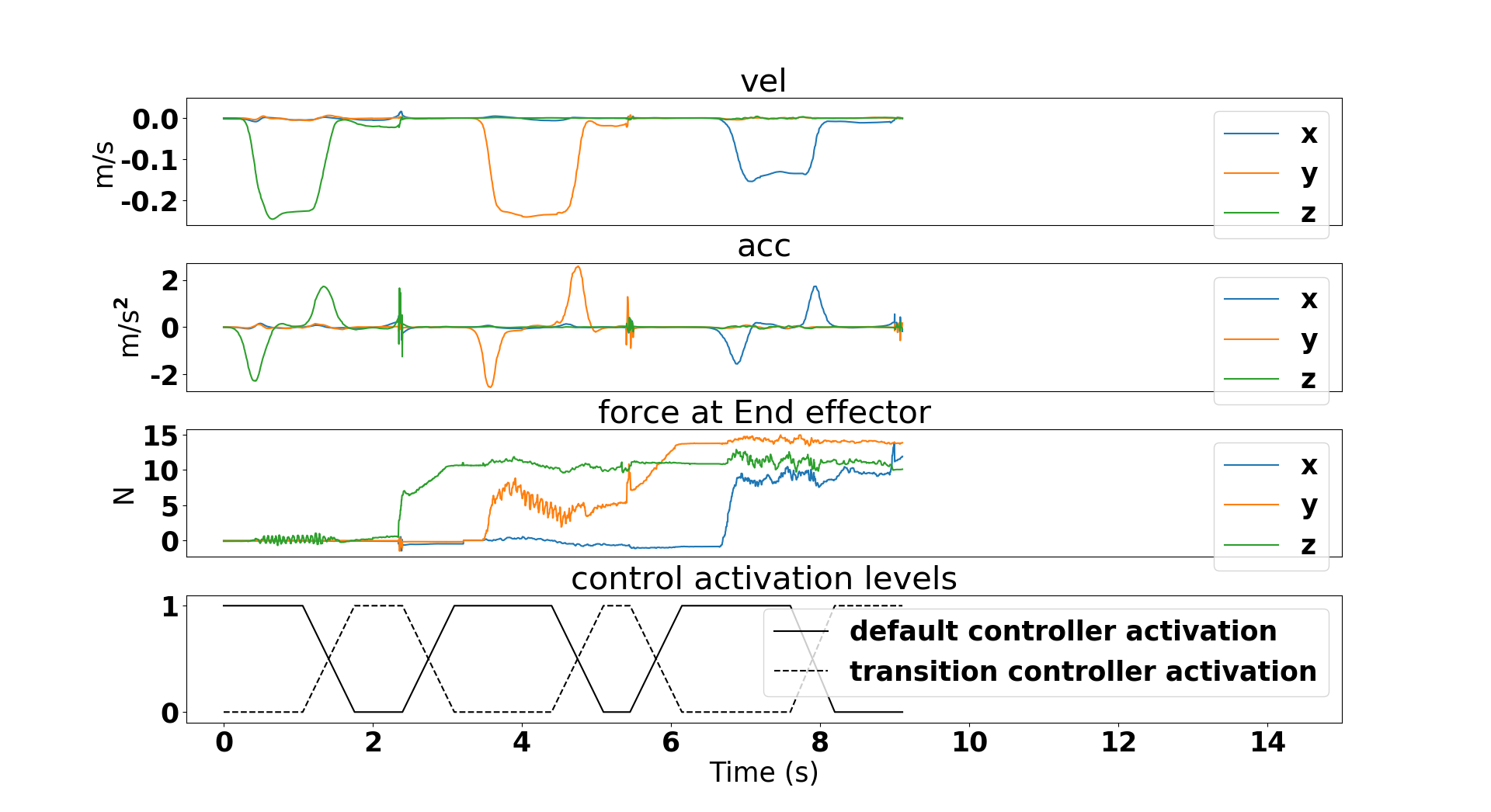}
      \vspace{-1em}
      \caption{Experiment trial 5.}
      \label{fig:expt_last_trial}
   \end{subfigure}
  \end{center}
  \vspace{-1em}
  \caption{Velocity, acceleration, force, and controller activation levels in: (a) Experimental trial 1; (b) Experimental trial 5. Use of our framework reduces uncertainty in estimates of contact positions, reduces the time spent using the transition-phase controller, and minimizes discontinuities.}
  \label{fig:exp_trials}
  \vspace{-1em}
\end{figure*}
  
To evaluate the overall framework and the resulting dynamics on a physical robot, the robot (with a wooden block attached to end-effector) was asked to move vertically down to the table (contact 1), slide along y-axis (on table surface) to a wall (contact 2), and slide along the wall (on table surface) to hit another obstacle (contact 3), as shown in Figure~\ref{fig:task_setup}. The robot was provided significantly wrong initial guesses of the contact positions with noise (see Table~\ref{tab:contact_preds}). The robot had to repeat the task while reducing the deviation from the given motion pattern by improving its belief about the positions of the contacts. The robot also had to modify its approach velocity from the initial value of $0.05\,m/s$ to produce a desired impact force of $8\,$N. Since each contact in the task is in the presence of different environment dynamics (e.g., motion in free space, motion against surface friction), the velocity required to attain the desired impact force was expected to be different. The robot also had to incrementally update its approach velocity for each contact using gradient descent till the desired velocity for that environment was achieved. The robot also had to perform all the trials with smooth overall motion dynamics with minimum spikes in the velocity or acceleration profiles.



Figure~\ref{fig:expt_first_trial} shows the velocity, acceleration, and EE force in the first trial, and Figure \ref{fig:expt_last_trial} shows these values after five trials. The results in these figures and in Table~\ref{tab:contact_preds} show that the uncertainty in the estimates of the contact positions is reduced, as indicated by a significant reduction in the size of the covariance ellipsoids, and the robot spends significantly less time using the transition-phase controller and the associated lower velocity. The activations of the default controller and the transition-phase controller are indicated in the last plot of these figures. The overall task could be completed in $9.2\,s$ in the fifth trial as opposed to $14.4\,s$ in the first trial. The covariance ellipsoids converged in the first three trials of the task, but the task was repeated to evaluate the ability to compute and set the approach velocity for different transition-phase controllers.

\begin{table}[tb]
\centering
\vspace{1.5em}
\begin{tabular}{lcr}
\toprule
 Prediction Error (m) & Initial & Final (trial 5)  \\
\midrule
Contact 1 (Z-axis) & 0.12 $\pm$ 0.3  & 0.016 $\pm$ 0.039 \\
Contact 2 (Y-axis) & 0.09 $\pm$ 0.2 & 0.011 $\pm$ 0.04 \\
Contact 3 (X-axis) & 0.1 $\pm$ 0.2 & 0.018 $\pm$ 0.036 \\
\bottomrule
\end{tabular}
\caption{Euclidean error in the estimated contact locations in the first and fifth trials of the task shown in Figure~\ref{fig:task_setup}. Values represent errors along the most significant axis for each contact (in parenthesis). The corresponding values along the diagonal of the covariance matrix are shown as the  standard deviation.}
\label{tab:contact_preds}
\vspace{-1.5em}
\end{table}
  
The framework converged to a suitable approach velocity for the first contact (from motion in free space) in five iterations. It was, however, difficult for the robot to adjust its approach velocities for contacts 2 and 3, which required the robot to use force control along one and two directions (respectively). Contact 3 was particularly challenging because it involved sliding along two different surfaces, resulting in very noisy readings from the force-torque sensor due to the different values of frictional resistance offered by the two surfaces. The impact force being along the same direction as friction also made it more difficult to isolate the impact force from the force due to surface friction.

\section{Discussion and Future Work}
\label{sec:discuss}
This paper described a framework towards addressing the discontinuities in changing-contact manipulation tasks. The framework introduces a transition-phase controller in a hybrid force-motion variable impedance controller for continuous contact tasks. Our representational choices enable us to simplify and address the associated challenges reliably and efficiently. Specifically, a Kalman filter-baed approach is used to incrementally improve the estimates of the contact positions. These estimates are used to minimize the time spent in the transition phase (with lower velocity and stiffness), and the velocity profile is modified automatically to smooth motion and a desired impact force. 

The framework opens up many directions of further research. First, we only focused on collisions due to translational motion, and did not address collisions due to rotations of the end-effector. This could be addressed by defining a region of anticipated collision in SO(3). Second, we observed that updating approach velocity for collisions when the robot is already in contact with another surface is more complicated. This is because of the difficulty in differentiating the sensor readings obtained due to reactive forces from the existing contact and the sensor readings obtained due to the impact force generated when colliding with another object. One possible way to address this issue is to learn a better forward model for the contact mode such that it can accurately predict the forces due to the first contact. Third, we only modified the velocity profile and position trajectory to achieve the desired smooth motion; future work will explore the relationship between stiffness values and the impact forces, and adapt the orientation as well. Initial experiments indicate that this is a challenging problem.  Furthermore, we could explore the use of other kinds of sensors (e.g., cameras) to provide additional information about contact positions. The overall objective is to eliminate or minimize discontinuities in changing-contact robot manipulation tasks.

%
%
%
\bibliographystyle{splncs04}
\bibliography{ref}

\end{document}